\documentclass[journal]{IEEEtran}
\usepackage{caption}
\usepackage{color}
\usepackage{comment}
\usepackage{colortbl}  %彩色表格需要加载的宏包
\usepackage{xcolor}
\usepackage{epsfig}
\usepackage{graphicx}
\usepackage{bbding}
\usepackage{makecell}
\usepackage[font=footnotesize]{caption}
\usepackage{array, threeparttable}

\usepackage{amsmath}
\usepackage{amssymb}
\usepackage{comment}
\usepackage{multirow}

\usepackage{hyperref}
\hypersetup{hidelinks}

\usepackage{booktabs}
\usepackage{subfigure}
\usepackage[ruled]{algorithm2e}
\usepackage{listings}
\usepackage{color} 
\usepackage{mathtools}

\usepackage{gensymb}
\usepackage{setspace}

\usepackage{algorithmic}
\usepackage[normalem]{ulem}
\usepackage{cancel}

\begin{document}

\title{\LARGE \bf IAD-GPT: Advancing Visual Knowledge in Multimodal Large Language Model for Industrial Anomaly Detection}

\author{Zewen Li, Zitong Yu, Qilang Ye, Weicheng Xie, Wei Zhuo and Linlin Shen
\thanks{
This work was supported by 
National Natural Science Foundation of China under Grant 62276170, 62306061, 62576076 and 82261138629, Open Fund of National Engineering Laboratory for Big Data System Computing Technology (Grant No. SZU-BDSC-OF2024-02), Guangdong Basic and Applied Basic Research Foundation (Grant No. 2023A1515140037), Guangdong-Macao Science and Technology Innovation Joint Fundation under Grant 2024A0505090003, Guangdong Provincial Key Laboratory under Grant 2023B1212060076, and Shenzhen Science and Technology Program (JCYJ20240813141807010). Corresponding authors: Zitong Yu (email: zitong.yu@ieee.org) and Linlin Shen (email: LLshen@szu.edu.cn).
}
\thanks{Z. Li, W. Xie and L. Shen are with School of Computer Science \& Software Engineering, Shenzhen University, China,
% Computer Vision Institute, School of Computer Science \& Software Engineering, Guangdong Key Laboratory of Intelligent Information Processing, Shenzhen University, Shenzhen
518060, and Z. Li is also with School of Computing and Information Technology, Great Bay University, Dongguan, 523000,China.}
\thanks{Z. Yu is with School of Computing and Information Technology, Great Bay University, Dongguan, 523000, China.}
\thanks{Q. Ye is with College of Computer Science, Nankai University, Tianjin.}
\thanks{W. Zhuo is with School of Artificial Intelligence, Shenzhen University, Shenzhen 518060, China, Guangdong Provincial Key Laboratory of Intelligent Information Processing, Shenzhen University, Shenzhen, China, and National Engineering Laboratory of Big Data System Computing Technology, Shenzhen University.}
}

\markboth{IEEE Transactions on Instrumentation and Measurement}%
{Shell \MakeLowercase{\textit{et al.}}: Bare Demo of IEEEtran.cls for IEEE Journals}

\maketitle

\begin{abstract}
The robust causal capability of Multimodal Large Language Models (MLLMs) hold the potential of detecting defective objects in Industrial Anomaly Detection (IAD). 
However, most traditional IAD methods lack the ability to provide multi-turn human-machine dialogues and detailed descriptions, such as the color of objects, the shape of an anomaly, or specific types of anomalies. At the same time, methods based on large pre-trained models have not fully stimulated the ability of large models in anomaly detection tasks.
In this paper, we explore the combination of rich text semantics with both image-level and pixel-level information from images and propose IAD-GPT, a novel paradigm based on MLLMs for IAD. We employ Abnormal Prompt Generator (APG) to generate detailed anomaly prompts for specific objects. 
These specific prompts from the large language model (LLM) are used to activate the detection and segmentation functions of the pre-trained visual-language model (i.e., CLIP).
% These prompts are used to activate the detection and segmentation capabilities of a pre-trained visual-language model, i.e. CLIP.
To enhance the visual grounding ability of MLLMs, we propose Text-Guided Enhancer, wherein image features interact with normal and abnormal text prompts to dynamically select enhancement pathways, which enables language models to focus on specific aspects of visual data, enhancing their ability to accurately interpret and respond to anomalies within images.
Moreover, we design a Multi-Mask Fusion module to incorporate mask as expert knowledge, which enhances the LLM's perception of pixel-level anomalies. Extensive experiments on MVTec-AD and VisA datasets demonstrate our state-of-the-art performance on self-supervised and few-shot anomaly detection and segmentation tasks, such as MVTec-AD and VisA datasets. The codes are available at \href{https://github.com/LiZeWen1225/IAD-GPT}{https://github.com/LiZeWen1225/IAD-GPT}.
\end{abstract}


\IEEEpeerreviewmaketitle
\vspace{-1em}
\section{Introduction}

\IEEEPARstart{T}{he} goal of IAD tasks is to identify defects in general objects that differ from normal patterns, such as scratches on leather, damaged capsules, etc. The application of anomaly detection in industry ensures the smooth progress of production processes and plays a crucial role in monitoring, maintaining, and optimizing industrial production processes.

\begin{center}
    \begin{figure}
        \centering
        \includegraphics[width=1\linewidth]{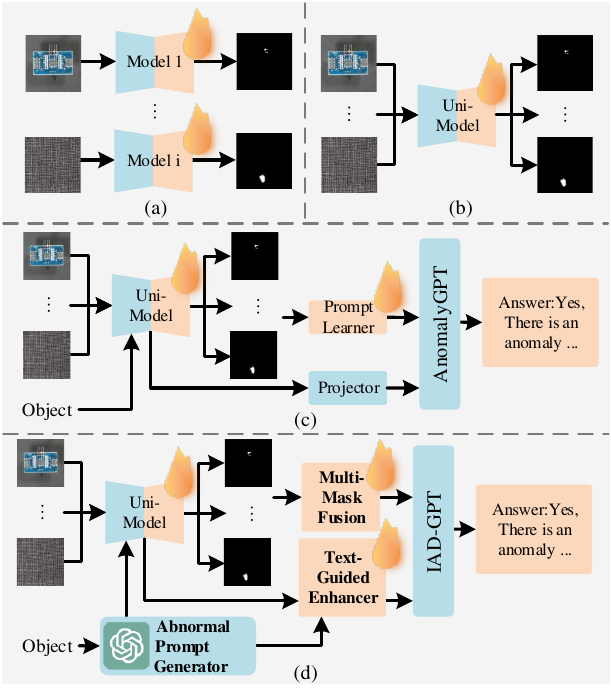}
        \caption{
        Comparison between our IAD-GPT, traditional IAD methods and AnomalyGPT.
        (a) Traditional methods use separate models for different classes and provide anomaly scores only.
        (b) Unified methods manage to accomplish anomaly detection for various classes with a unified framework.
        (c) AnomalyGPT, based on the settings in (b), enhances the pixel-level visual knowledge of MLLMs to perceive anomalies.
        % AnomalyGPT, based on the settings in (b), enhance pixel-level visual knowledge of MLLMs is used to perceive anomalies.
        (d) IAD-GPT provides GPT-generated abnormal text to improve localization capabilities and enhances image-level and pixel-level visual knowledge to achieve better anomaly recognition by MLLMs.
        % IAD-GPT, provide GPT-generated abnormal text to improve localization capabilities, and enhance image-level and pixel-level visual knowledge to achieve better recognition of anomalies by MLLMs.
        }
        % \vspace{-30pt}
        \label{fig:fig1}
    \end{figure}
    % \vspace{-25pt}
\end{center}

\vspace{-25pt}
The research on IAD tasks \cite{memorizing, gu2024anomalygpt, simplenet, ristea2022self, patchcore, uniad, realnet} is constantly developing and making good progress. Current mainstream methods \cite{gu2024anomalygpt, simplenet, patchcore, uniad, realnet} for IAD include feature embedding-based methods \cite{gu2024anomalygpt, simplenet, patchcore} and reconstruction-based methods\cite{uniad, realnet}. However, traditional IAD methods are currently limited to providing anomaly detection and segmentation results for objects. These approaches all rely on manually setting thresholds and lack the capability to offer detailed insights into the nature and specifics of detected defects.
Meanwhile, image-text matching is often used to detect anomalies in large pre-trained model-based approaches like WinCLIP \cite{jeong2023winclip}, which uses a compositional prompt ensemble based on text templates using generic descriptions of normal/ abnormal as text. This has been followed by other researchers in subsequent studies \cite{gu2024anomalygpt,cao2024adaclip,zhouanomalyclip}, but the method does not fully activate the capability of the large pre-trained model. Filo \cite{gu2024filo, gu2025filo++} proposes an adaptively learned Fine-Grained Description that leverages domain-specific knowledge to introduce detailed anomaly descriptions, replacing generic normal and abnormal descriptions.

Research progress on LLMs has been rapid recently. Due to their excellent language understanding and reasoning abilities after large-scale data training, LLMs such as ChatGPT \cite{chatgpt} and Llama \cite{llama} have proven their ability to perform translation, paraphrasing, and instruction following tasks in zero sample tasks. In the research of MLLMs \cite{otter, llava, minigpt}, it is found that other modal information can be mapped to the feature space of LLMs through fine-tuning. LLM can also understand the information contained in other modalities and make explanations for it.
% AnomalyGPT \cite{gu2024anomalygpt} utilizes the powerful capabilities of LLM in areas such as instruction following, answering, and dialogue.
AnomalyGPT \cite{gu2024anomalygpt} is the first to introduce LLMs into IAD and proposes the task of Anomaly Perception in Multimodal Large Language Models (APMLLM).
% The introduction of LLMs into industrial anomaly detection tasks for the first time has achieved excellent results.
MLLMs for anomaly detection eliminate the problem of manually setting thresholds in traditional methods and make the results of industrial anomaly detection and localization more interpretable. However, AnomalyGPT simply fine-tunes image features into LLM through a linear layer, feeds predicted masks as expert knowledge into LLM, and finally allows LLM to make judgments on image anomalies.
% However, AnomalyGPT has not fully stimulated the ability of LLMs and pre-trained visual-language models in industrial anomaly detection tasks. 

In this paper, we propose IAD-GPT, which is designed to enhance the efficiency and accuracy of anomaly detection in industrial quality inspection. This method not only supports multi-turn human-machine dialogues, allowing operators to delve into potential anomalies through interactive question-and-answer (QA) sessions, but also leverages advanced LLMs to directly analyze anomalies within images without relying on pre-set threshold values for anomaly detection. Traditional anomaly detection methods typically employ fixed threshold standards: If the detected anomaly value exceeds a certain threshold, the image is flagged as containing an anomaly; otherwise, it is considered normal. In contrast, our approach offers greater flexibility and adaptability by making precise judgments based on specific contexts and using LLMs to directly output intuitive results. Consequently, this method holds significant potential for practical application in production environments, providing a novel perspective and solution for industrial quality inspection.

Fig.~\ref{fig:fig1} shows the difference between our IAD-GPT and previous research. To address the issue of insufficient stimulation of large pre-trained model segmentation ability in the compositional prompt ensemble method \cite{jeong2023winclip}. we employ APG to extend and enrich the semantic content of text prompts. These prompts are used to activate the detection and segmentation capabilities of a pre-trained visual-language model, i.e., CLIP \cite{clip}. Specifically, we leverage GPT's existing knowledge of most objects in the text domain and use a QA format to generate possible anomaly categories for each object class. These generated texts will serve as one of the key factors in identifying anomalies.
To enable the LLM to fully perceive image information, we designed two modules at the image level and pixel level, respectively: Text-Guided Enhancer (TGE) and Multi-Mask Fusion (MMF). TGE enhances the LLM's anomaly perception capability at the image level by interacting image features with normal/abnormal text prompts to achieve dynamic path selection. Meanwhile, the MMF uses the differences in image-text features across multiple levels to further improve the LLM's anomaly perception capability at pixel level.

% In this paper, we base on the existing multimodal large language model (MLLM), and obtain the specific anomalies about each different category of images as a priori knowledge from the powerful LLM through the form of QA session, and then stimulate the capability of the large model to achieve a better detection and segmentation performance.

% The method proposed in this paper is called IAD-GPT, we fully leverage the powerful ability of LLM to universally understand things and stimulate pre-trained visual-language models to develop a more comprehensive understanding of abnormal patterns and defects. Furthermore, we introduce the AnomalyMoE module and the Multi-Mask Fusion module to enhance visual features, facilitating their better alignment with the feature space of LLM. This alignment significantly improves LLM's ability to recognize anomalies, enabling it to deliver more accurate and detailed responses.

Our contributions are summarized as follows:
\begin{itemize}
\item We introduce a novel framework named IAD-GPT, via leveraging rich visual knowledge for IAD. Compared with previous IAD methods, IAD-GPT enhances the capability to perceive anomalies beyond traditional approaches.
\item  We employ APG to generate detailed anomaly prompts for specific objects.
These prompts are utilized to activate the detection and segmentation capabilities of pre-trained visual-language models via incorporating rich semantic information can significantly enhance the performance of large pre-trained models in IAD tasks.
\item For the task of APMLLM, we design a multi-scale feature enhancement approach. At image level, we develop TGE to dynamically select enhancement paths for image features. At pixel level, we introduce MMF, which leverages differences in image-text features across multiple levels to improve the LLM's ability to perceive the location of anomalies.
% We design of two modules, Text-Guided Enhancer and Multi-Mask Fusion, at the image level and pixel level respectively, to enhance the LLM's perception of image information. The Text-Guided Enhancer improves anomaly detection at the image level through dynamic path selection by interacting image features with normal/abnormal text prompts. Meanwhile, Multi-Mask Fusion enhances pixel-level anomaly perception by using differences in image-text features across multiple levels.
\item We achieve state-of-the-art performance on MVTec-AD and VisA for self-supervised/few-shot anomaly detection and segmentation tasks. Compared to the baselines, IAD-GPT shows superior performance in anomaly detection and localization on images within a self-supervised learning setting, outperforming the few-shot setting.
\end{itemize}

The remainder of this paper is organized as follows. Section II reviews the related works. In Section III, we describe the proposed approach in detail. Section IV presents ablation studies and comparison experiments with state-of-the-art methods. Finally, Section V provides conclusions and outlines directions for future work.

\begin{figure*}
\centering
\includegraphics[scale=0.5]{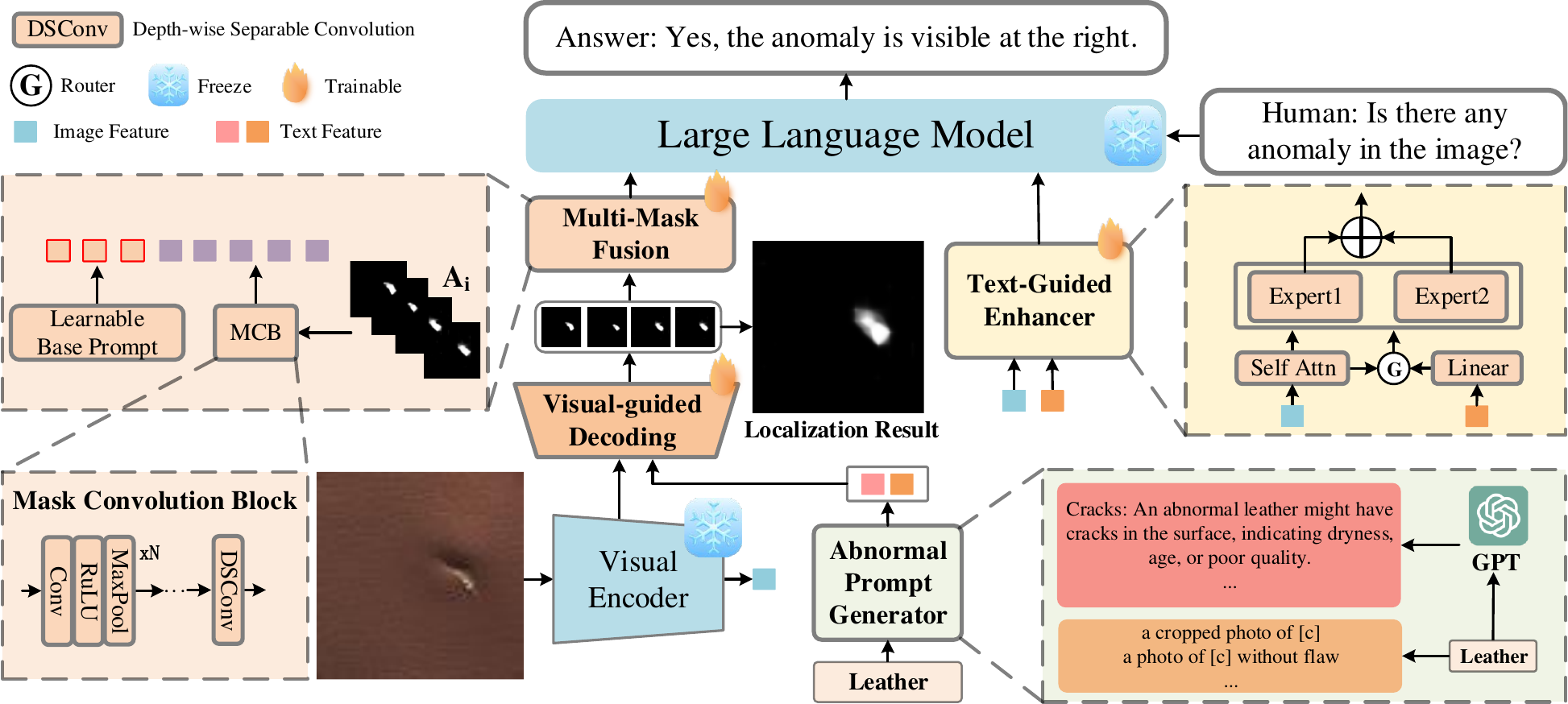}
\caption{
  Overview of IAD-GPT. Abnormal Prompt Generator provides category specific text prompts for decoder. Text-Guided Enhancer and Multi-Mask Fusion provide image-level visual information and pixel-level expert knowledge to LLMs, respectively.
  }

\label{fig:model}
\vspace{-1.1em}
\end{figure*}

% \vspace{-10pt}
\section{Related Work}
\label{sec:relatedwork}

\subsection{Industrial Anomaly Detection}
Industrial anomaly detection is mainly divided into reconstruction-based methods and feature embedding-based methods.

Reconstruction-based methods \cite{memorizing, ristea2022self, uniad, realnet, zavrtanik2021draem, cao2025varad, zavrtanik2021reconstruction} rely on using only normal data when training the model, learning the feature distribution of normal data to reconstruct normal features. In the test phase, the trained model reconstructs the query data to obtain the normal feature of the query image and then compares the difference between the reconstructed image features and the original query image features to achieve the detection and location of anomalies. RealNet \cite{realnet} uses a diffusion model with controllable strength to synthesize abnormal data and training the reconstruction network with abnormal data that are more similar to real-world anomalies.

Feature embedding-based \cite{simplenet, patchcore,liang2025tocoad,xie2025semi,padim,timzhai2025bidirectional,timzhang2025edgead,timzhou2024dual} methods often use networks trained on ImageNet \cite{imagenet} to extract features from images. The representation of abnormal areas in the image's feature space is usually far away from normal feature clusters, and anomaly detection is achieved through the obvious distance between them in the feature space. For example, PatchCore \cite{patchcore} constructs a memory bank storing representative patch-features from normal images to detect anomalies in industrial settings without needing any examples of defects. It employs locally aware patch features aggregated from intermediate feature hierarchies of a pre-trained network, ensuring spatial resolution and generality. To manage the size of the memory bank and maintain performance, PatchCore applies a coreset sub-sampling mechanism that selects a subset of features for efficient nearest neighbour computations. However, networks pre-trained on general datasets often lack expertise in the field of IAD. Migrating general pre-trained networks to specific IAD downstream tasks can make the model perform better. In SimpleNet \cite{simplenet}, a two-layer adapter is used to transfer the features extracted from the pre-trained network to the domain, synthesize abnormal data in the feature space, and then train a simple discriminator to achieve excellent anomaly detection results.

Previous studies on IAD has mainly focused on ``one model for one class", and there is little research on unified anomaly detection models. UniAD \cite{uniad} is a model specifically used for unified anomaly detection. UniAD uses learnable query and neighbor masking attention to prevent the model from taking shortcuts, thereby building a more robust unified anomaly detection model. DiAD \cite{he2024diffusion} leverages advanced diffusion models to enhance the reconstruction and localization of anomalies across various classes. By incorporating learnable query and neighbor masking attention mechanisms in UniAD, DiAD prevents shortcut learning, thereby building a more robust model.
With the powerful capabilities of pre-trained visual-language models such as CLIP, unified IAD models have more research potential. WinCLIP \cite{jeong2023winclip} uses the characteristics of CLIP to align images and texts, and uses CLIP for IAD tasks. Specifically, WinCLIP calculates the similarity between multi-scale image features and text features representing normal/abnormal features, thereby realizing the detection of abnormal areas. AnomalyGPT \cite{gu2024anomalygpt} uses ImageBind \cite{imagebind} to train a simple decoder to align the feature space of images and texts to achieve industrial anomaly detection. By employing generalized object-agnostic text prompt templates, AnomalyCLIP \cite{zhouanomalyclip} learns embeddings for normality and abnormality, further enhanced by global and local context optimizations to better understand anomaly semantics. AdaCLIP \cite{cao2024adaclip} enhances the performance of the CLIP model in zero-shot anomaly detection (ZSAD) by utilizing hybrid learnable prompts, and emphasizes the importance of optimizing cues for detecting anomalies in individual images. FiLo \cite{gu2024filo} enhances the perception of anomalies in ZSAD tasks through Fine-Grained Descriptions and high-quality localization with Position Enhancement. 

Previous studies utilize the powerful capabilities of large pre-trained models, but do not fully stimulate the large pre-trained models to locate anomalies at the pixel level. In this paper, based on LLM and the prior knowledge of the large pre-trained model, we generate possible abnormal attributes for the categories that may be encountered in the unified anomaly detection process. Specific prompts fully stimulate the capabilities of the large pre-trained model, and we have achieved excellent results.

% \vspace{-10pt}
\subsection{Multimodal Large Language Model}
With the significant progress of LLMs like ChatGPT and GPT-4 \cite{chatgpt}, many studies have attempted to explore other modes based on LLMs, connecting pre-trained visual-language models of different modalities into end-to-end trainable models, also known as multimodal large language models. Due to the excellent language understanding and reasoning abilities of LLMs after large-scale data training, such as Qwen \cite{bai2023qwen} and Llama \cite{llama}, They have demonstrated their ability to perform translation, paraphrasing, and instruction following tasks in zero reference tasks.
Models such as MiniGPT-4 \cite{minigpt}, Llava \cite{llava}, and InstructBLIP \cite{Instructblip} all employ fine-tuning techniques~\cite{hu2022lora,loraadapter,faceanti-spoofing1} to construct MLLMs.
\mbox{MiniGPT-4}~\cite{minigpt} uses frozen Qformer and image encoder for image feature extraction based on BLIP2 \cite{li2023blip}, and trains a simple linear layer to align visual modalities into the LLM. Llava \cite{llava} is similar in architecture to MiniGPT-4, but through more diverse data and fine-tuning strategies at different stages, Llava is able to complete more complex reasoning. InstructBLIP \cite{Instructblip} has conducted a comprehensive and systematic study on the fine-tuning of visual language instructions. The InstructBLIP model benefits from adopting a balanced sampling strategy to synchronize learning progress across datasets, enabling it to achieve excellent zero sample performance on various visual language tasks. The above-mentioned multimodal big language models mainly use visual encoders pre-trained on roughly aligned image text pairs, resulting in insufficient extraction and inference of visual knowledge. 
Therefore, more research work~\cite{otter,faceanti-spoofing2,xie2024fusionmamba,chen2024lion,llamaadapter1,ye2025cat} related to multimodal alignment is proposed.
To address this issue, LION \cite{chen2024lion} designed a multi granularity fusion visual aggregator and used image labels as advanced semantic visual information, enabling LION to have more advanced overall and fine-grained visual perception capabilities. In addition to providing visual information as input only to LLM, method such as LLaMA Adapter \cite{llamaadapter2, llamaadapter1}, Multi modal GPT, and Otter \cite{otter} also fuse multimodal information with intermediate features in LLMs to achieve the understanding of multimodal information by LLMs. CAT~\cite{ye2025cat,cat+} designed a clue aggregator to aggregate clues related to problems in dynamic audio-visual scenes, targeting rich and complex dynamic audio-visual compositions. This enriches the detailed knowledge required for learning, enabling CAT to learn clues related to problems and directly engage in action based audio-visual reasoning. CAT outperforms other MLLMs in multimodal tasks, especially audio-visual question answering tasks.

MLLMs are trained on large-scale general datasets, which limits their capability to specifically perceive anomalies. To overcome this challenge, we introduce a method that utilizes image-level visual information and pixel-level expert knowledge. By integrating these rich sources of information, our approach significantly enhances the ability of MLLMs to perceive anomalies, thereby improving their performance in the APMLLM task.

% MLLMs lack the ability to perceive anomalies because they are trained on large-scale general datasets. We propose to use image level visual information from multiple scales and pixel level expert knowledge to provide richer information for multimodal large language model to perceive anomalies.
 
% In this article, we designed a Mixture of Experts (MoE) module called AnomalyMoE to align the feature spaces of LLM and pre-trained image encoder, which enable LLM to better understand image-level anomaly.

\section{Methodology}
Fig.~\ref{fig:model} illustrates the architecture of IAD-GPT. Given a query image $x \in \mathbb{R}^{H\times W \times C}$, the visual features $F_{img} \in \mathbb{R}^{1 \times C_1}$ extracted by the image encoder are passed through TGE to obtain the image embedding $E_{img} \in \mathbb{R}^{1 \times C_{emb}}$, which is then fed into the LLM.

Our method is experimentally validated in two distinct settings: an self-supervised setting, where the model learns from data with only normal samples, and a few-shot setting, which challenges the model to generalize from a very limited number of normal samples.
In self-supervised setting, the patch-level features extracted by intermediate layers of image encoder are fed into the decoder together with text features that expand anomaly prompts with APG to generate pixel-level anomaly localization results.
In few-shot setting, the patch-level features from normal samples are stored in memory banks and the localization result can be obtained by calculating the distance between query patches and their most similar counterparts in the memory bank. 
The localization result is subsequently transformed into prompt embeddings $E_{fusion} \in \mathbb{R}^{L_1 \times C_{emb}}$ through the MMF module, serving as a part of the LLM input. The LLM detects anomalies and identifies their locations by leveraging the image input $E_{img}$, prompt embedding $E_{fusion}$, and user-provided text, thereby generating a response for the user.
\begin{center}
    \begin{figure}
        \centering
        \includegraphics[scale=0.8]{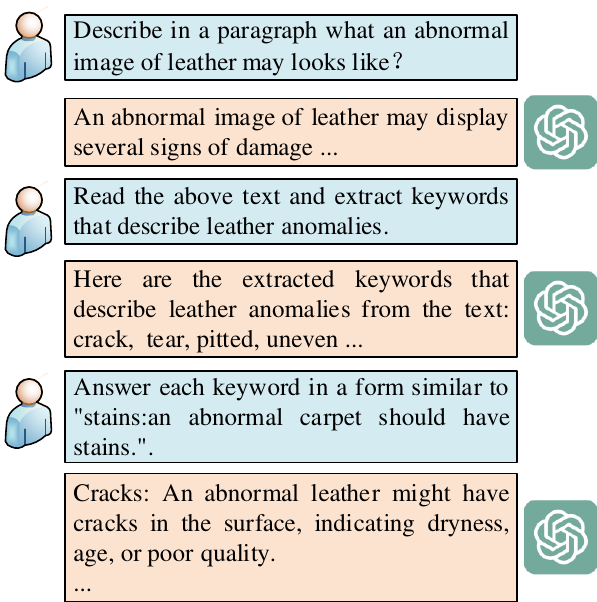}
        % width=1\linewidth
        \caption{Example of APG for leather. We improve the stability of LLM-generated prompts by designing an QA session and providing illustrative examples.
        }
        \vspace{-15pt}
        \label{fig:APG}
    \end{figure}
    \vspace{-25pt}
\end{center}

% \vspace{-10pt}
\subsection{Abnormal Prompt Generator}
We design Abnormal Prompt Generator (APG) to expand anomaly prompts to achieve more powerful segmentation capabilities. Specifically,we first prompt the LLM with the query: ``\textit{Describe in a paragraph what an abnormal image of \{$C_o$\} may looks like?}" with the given class $C_o$. And we extract potential abnormal attributes $ATTR_a$ from the answer generated by LLM. $ATTR_a  = \left\{k_1,k_2,...,k_i\right\} $ includes several potential abnormal keywords $k_i$ for $C_o$. For each potential abnormal keyword $k_i$, we continue the QA session to generate an class-keyword abnormal prompt $T_{k_i}$. $T_{apg}=\left\{T_{k_1},T_{k_2},...,T_{k_i}\right\}$ contain all potential class-keyword abnormal prompts generated by multiple rounds of dialogue. For example, for leather objects, LLM is used to answer the relevant abnormal categories, including Irregular texture, Tears, Cracks, etc. Then LLM is asked to generate corresponding text for each abnormal category, such as ``\textit{Cracks: An abnormal leather may have cracks in the surface, indicating dryness, age, or poor quality.}". 
\begin{figure*}
\centering
\includegraphics[scale=0.5]{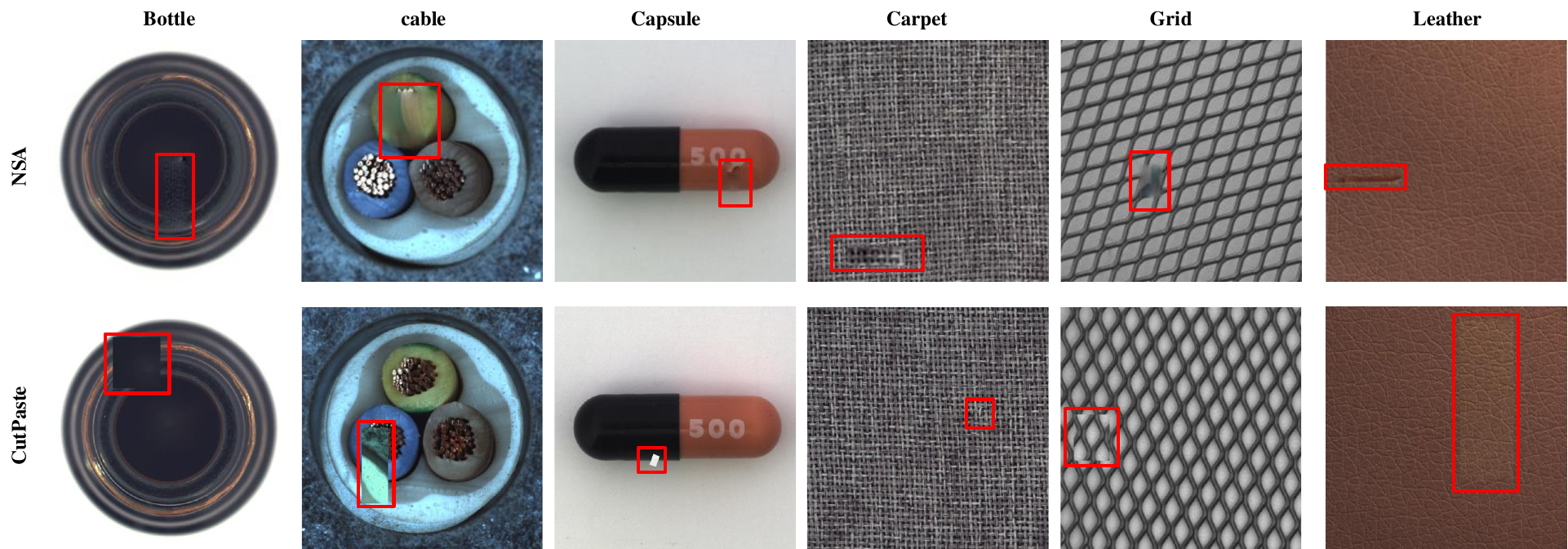}
% \captionsetup{justification=justified, singlelinecheck=false}
\caption{
  Visualization comparison of anomaly image generation results between NSA and CutPaste methods. Red box indicates abnormal area.
  }
\label{fig:NSA}
\vspace{-1.1em}
\end{figure*}
Fig.~\ref{fig:APG} shows the process of using APG to generate specific anomaly categories for leather and converting them into text prompts. We not only simply expand the anomaly categories into fixed format text, but also let the LLM infer the characteristics of the object based on its own and generate appropriate text prompts.
WinCLIP \cite{jeong2023winclip} introduces a two-class design method in text prompts to help CLIP locate the abnormal region, which categorizes the text prompts into normal prompts $T_n$ and abnormal prompts $T_a$, we define text prompts similar to WinCLIP as $T_{win} =\{T_n, T_a\}$.
When training the image decoder, we use $T_{win}$ and $T_{apg}$ as our text prompts $T_{text}=\{T_{win}, T_{apg}\}$,
We then extract the text embeddings $F_{text} \in \mathbb{R}^{L_2 \times C_2}$ using the pre-trained CLIP model and align the patch-level image features $F_{patch} \in \mathbb{R}^{H \times W \times C_3}$ with the text embeddings $F_{text}$ through a simple linear layer.
The anomaly score is calculated by the similarity between the patch feature $F_{patch}$ and text embeddings $F_{text}$. The localization result $M \in \mathbb{R}^{H \times W}$ can be obtained as follow:
\begin{center}
\vspace{-10pt}
\begin{equation}
M = Unsample\left(\sum^{4}_{l=1} Softmax\left(Linear\left(F^{l}_{patch}\right)F_{text}^{T}\right)\right)
\end{equation}
\vspace{-10pt}
\end{center}
where $l$ represents the number of layers. Similar to AnomalyGPT, we do not specifically select image features of different layers for mask generation. The reason is that image features have different effects on anomaly extraction in shallow and deep layers. In previous studies, it has been found that fusing shallow and deep features helps us generate masks more accurately. For multi-layer masks, we sum them up and calculate the average to obtain the final predicted mask, which is then achieved through upsampling.

For few-shot IAD, we utilize the same image encoder to extract patch-level features from normal samples and store them in memory banks $B^l \in \mathbb{R}^{N \times C_3}$. For patch-level features $F^{l}_{patch} \in \mathbb{R}^{H \times W \times C_3}$. The IAD localization results under the few-shot setting can be expressed as follows:
\begin{center}
\vspace{-10pt}
\begin{equation}
M = Unsample\left(\sum^{4}_{l=1}\left( 1 - Max\left(F^{l}_{patch} \cdot B^{l^{T}}\right)\right)\right)
\end{equation}
\vspace{-15pt}
\end{center}

\subsection{Text-Guided Enhancer}
PandaGPT \cite{pandagpt} uses a simple linear layer to align the feature space of the image encoder and LLM. However, PandaGPT has not been trained for data in the field of industrial anomalies, resulting in PandaGPT being unable to identify anomalies during industrial anomaly detection.
Inspired by the Mixture-of-Experts (MoE) architecture \cite{moe}, we propose the Text Guided Enhancer (TGE) module, in which a similar structure is designed to enhance image-level features. However, unlike the traditional MoE approach that employs a Router, we dynamically control feature enhancement for each individual image through the interaction between $F_{img}$ and $F_{win}$.

% To enhance the MLLM's perception of image-level anomaly, we propose Text-Guided Enhancer (TGE), wherein image features $F_{img} \in \mathbb{R}^{1 \times C_1}$ interact with normal or abnormal text prompts to dynamically select enhancement pathways. This approach enables LLMs to focus on specific aspects of visual data, enhancing their ability to accurately interpret and respond to anomalies within images.
\begin{center}
\vspace{-15pt}
\begin{equation}
W_e = Softmax(Attn(F_{img}) Linear(F_{win})^{T})
\end{equation}
\vspace{-15pt}
\end{center}
where $F_{win}$ is the text embedding extracted by text encoder from $T_{win}$. We use a linear layer to align $F_{win}$ and image-level image feature $F_{img}$. The enhanced image-level feature of $F_{img}$ after self-attention is used as expert input, $W_e$ is used as the weight of expert aggregation, and the result $E_{img} \in \mathbb{R}^{1 \times C_{emb}}$ after expert aggregation is fed to LLM as image-level feature input.
\begin{center}
\vspace{-15pt}
\begin{equation}
E_{img} = \sum_{i=0}^{L_2}{W_{e_{i}} \times Expert_i(F_{img})}
\end{equation}
\vspace{-10pt}
\end{center}
where $L_2$ indicates the number of categories of $T_{win}$, and $Expert_i$ denotes the $i$-th expert. Our experts are composed of a combination of an attention block and a feed-forward neural network.

\begin{center}
    \begin{figure}
        % \centering
        \includegraphics[width=1\linewidth]{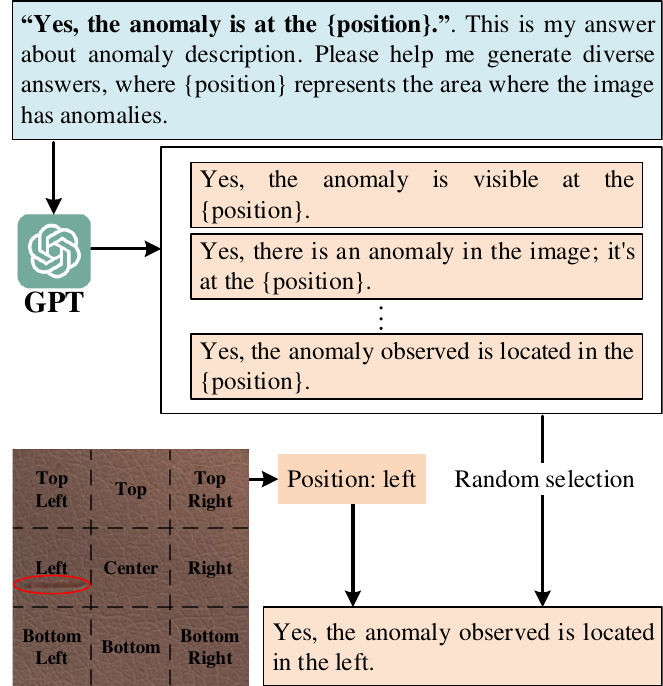}
        \caption{Illustration of generating abnormal prompts and a 3x3 grid of images for LLM to answer abnormal locations. We first input the answer template into LLM to generate diversified answers to improve the diversity and stability of model training. Then randomly select a template and fill the location information of the generated abnormal image into the answer.
        }
        \vspace{-10pt}
        \label{fig:dataforllm}
    \end{figure}
\end{center}

\vspace{-30pt}
\subsection{Multi-Mask Fusion}
To utilize the masks generated by the decoder as expert knowledge and maintain semantic consistency between the LLM and the decoder output, we introduce a Multi-Mask Fusion (MMF) method, which converts the localization results $M_i \, (i=1,2,3,4)$ into a prompt embedding $E_{fusion}$. As shown in the left side of Fig.~\ref{fig:model}, MMF consists of multiple convolutional neural networks and trainable base prompt embeddings $E_{base} \in \mathbb{R}^{L_3 \times C_{emb}}$.
Our convolutional neural network is designed to consist of multiple general convolutional layers, followed by depthwise separable convolutions. We refer to this network as Mask Convolution Block (MCB). The MCB converts localization result $M_i$ into prompt embeddings $E_{dec_i} \in \mathbb{R}^{L_1 \times C_3}$, and then concatenates multiple $E_{dec_i}$ in the channel dimension to obtain an embedding $E_{fusion} \in \mathbb{R}^{L_1 \times C_{emb}}$ that fuses multi-layer information. Expert knowledge $E_{expert} = \{E_{fusion},E_{base} \}\in \mathbb{R}^{(L_1+L_3) \times C_{emb}}$ concatenates $E_{base}$ and $E_{fusion} $ in the length dimension and ultimately inputs them into the LLM.
\begin{center}
\vspace{-15pt}
\begin{equation}
E_{dec_i} = MCB\left(M_i\right)
\end{equation}
\vspace{-20pt}
\end{center}

\begin{center}
\vspace{-10pt}
\begin{equation}
E_{fusion} = Concat\left(\{E_{dec_i}\}^4_{i=1}\right)
\end{equation}
\vspace{-20pt}
\end{center}

\subsection{Data for Training}
We use the NSA \cite{NSA} method for training. The NSA method advances the CutPaste \cite{li2021cutpaste} technique by integrating the Poisson image editing \cite{perez2023poisson} approach to mitigate the discontinuity caused by pasting image segments. In the domain of IAD, the CutPaste \cite{li2021cutpaste} is a prevalent method used for generating simulated anomaly images. This approach involves randomly cropping a block region from an image and pasting it onto a random location within the same or another image, thereby creating a simulated anomalous portion. While this method significantly enhances the performance of IAD models, it often results in noticeable discontinuities due to the abrupt insertion. To address these visual inconsistencies, the Poisson image editing method \cite{perez2023poisson} seamlessly integrates an object from one image into another by solving Poisson partial differential equations, thereby reducing visible artifacts from direct pasting. 
% Fig.~\ref{fig:NSA} shows an example generated by the Poisson image editing method.
Fig.~\ref{fig:NSA} presents a visual comparison between the image results generated by the NSA method and the original CutPaste method, clearly illustrating the improvement of the NSA method in mitigating discontinuities.

In order to prevent overfitting of LLM, we use the LLM to enrich our target prompt before training LLM. For normal images, our response is designed as \textit{``No, there are no abnormalities in the image."}. For abnormal images, we first generate different answer templates through LLM and define the position information of anomalies as $position$. Every time training data are generated, one of the answer templates will be selected and the position will be filled in as the answer, such as \textit{``Yes, the anomaly is visible at \{$position$\}."} or \textit{``Yes, there is an anomaly in the image; it's at the \{$position$\}."}, etc. For position information of anomalies $position$, we divide the image into a grid of $3 \times 3$ distinct regions to facilitate the LLM to answer the positions of anomalies, as shown in Fig.~\ref{fig:dataforllm}.

\begin{figure}
\centering
\includegraphics[scale=1.1]{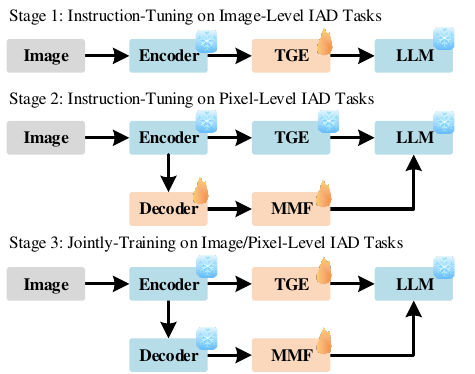}
% \vspace{-1.8em}
\caption{
  Training strategy of IAD-GPT.
  }
\label{fig:training}
\end{figure}

\subsection{Loss Functions}
To train our IAD-GPT, we primarily employed three loss functions: cross-entropy loss, focal loss \cite{focalloss}, and dice loss~\cite{diceloss}. The latter two are primarily utilized to enhance the pixel-level localization accuracy of the decoder. We only use cross-entropy loss when not training the decoder. And use all three losses when training the decoder.

\textbf{Cross-Entropy Loss} is a widely used loss function for training classification models. It quantifies the difference between the predicted probability distribution and the true distribution (often represented as one-hot encoded labels). The large language model is trained with cross-entropy loss, which quantifies the difference between the text sequence generated by the model and the target text sequence.
\begin{center}
\vspace{-10pt}
\begin{equation}
L_{c} = -\sum_{i=1}^{n} y_ilog(p_i)
\end{equation}
\vspace{-10pt}
\end{center}
where $n$ is the number of tokens, $y_i$ is the true label for token i and $p_i$ is the predicted probability for token $i$.

\textbf{Focal Loss} is an optimized loss function specifically tailored for addressing class imbalance in classification tasks, particularly within the realm of object detection. 
\begin{table*}[htbp]
\begin{center}
\caption{
QUANTITATIVE RESULTS (IMAGE-LEVEL AUROC/PIXEL-LEVEL AUROC/ACCURACY) OF SELF-SUPERVISED ANOMALY DETECTION TASKS ON MVTEC-AD DATASET. WE USE \textbf{BOLD} AND \underline{UNDERLINE} IN THE AVERAGE INDEX TO INDICATE THE BEST AND SUBOPTIMAL RESULTS RESPECTIVELY.
}
\setlength{\tabcolsep}{1.0mm}{
\begin{tabular}{c c c c c c c c}
\hline

\hline
Method/&Draem &PatchCore&SimpleNet&UniAD&DiAD&AnomalyGPT&IAD-GPT \\
% \cline{2-8} 
 Category& \text{(ICCV 21)}& \text{(CVPR 22)}& \text{(CVPR 23)}&\text{(NeurIPS 22)} &\text{(AAAI 24)}& \text{(AAAI 23)}& \textbf{(Ours)}\\
\hline
 \text{Bottle}& 97.5/87.6/- & 100/97.4/- & 97.7/91.2/-& 100/96.4/-& 99.7/98.4/-& 99.6/94.5/97.6 & 99.8/98.0/100
 \\
 \text{Cable}& 57.8/71.3/-& 95.3/93.6/-& 87.6/88.1/-& 95.2/97.3/-& 94.8/96.8/-& 89.8/86.4/83.3 & 93.6/91.4/88.7
 \\
 \text{Capsule}& 65.3/50.5/-& 96.8/98.0/-& 78.3/89.7/-& 86.9/98.5/-& 89.0/97.1/-& 95.1/93.2/87.9 & 97.8/98.3/94.7
 \\
 \text{Hazelnut}& 93.7/96.9/-& 99.3/97.6/-& 99.2/95.7/-& 99.8/98.1/-& 99.5/98.3/-& 99.1/91.9/94.5  & 100/98.7/97.3
 \\
 \text{Metal nut}& 72.8/62.2/-& 99.1/96.3/-& 85.1/90.9/-& 99.2/94.8/-& 99.1/97.3/-& 100/94.6/100 & 100/98.7/100
 \\
 \text{Pill}& 82.2/94.4/-& 86.4/90.8/-& 78.3/89.7/-& 93.7/95.0/-& 95.7/95.7/-& 94.8/84.4/88.0 & 94.7/97.8/88.7
 \\
 \text{Screw}& 92.0/95.5/-& 94.2/98.9/-& 45.5/93.7/-& 87.5/98.3/-& 90.7/97.9/-& 90.2/97.3/80.6 & 95.4/98.9/90.0
 \\
 \text{Toothbrush}& 90.6/97.7/-& 100/98.8/-& 94.7/97.5/-& 94.2/98.4/-& 99.7/99.0/-& 98.6/98.2/95.2 & 97.8/98.6/95.2
 \\
 \text{Transistor}& 74.8/64.5/-& 98.9/92.3/-& 82.0/86.0/-& 99.8/97.9/-& 99.8/95.1/-& 96.8/75.0/92.0 & 88.0/85.7/79.0
 \\
 \text{Zipper}& 98.8/98.3/-& 97.1/95.7/-& 99.1/97.0/-& 95.8/96.8/-& 95.1/96.2/-& 99.3/96.4/88.7 & 98.4/99.0/98.0
 \\
\hline
Object avg.& 82.6/81.9/-& 96.7/95.9/-& 84.8/92.0 /-& 95.2/\textbf{97.2}/-& \underline{96.3}/\textbf{97.2}/- & \underline{96.3}/91.2/\underline{90.8}& \textbf{96.5}/\underline{96.5}/\textbf{93.2}
\\
\hline
 \text{Carpet}& 98.0/98.6/-& 97.0/98.1/-& 95.9/92.4/-& 99.8/98.5/-& 99.4/98.6/-& 100/99.4/98.3 & 100/99.5/100
 \\
 \text{Grid}& 99.3/98.7/-& 91.4/98.4/-& 49.8/46.7/-& 98.2/96.5/-& 98.5/96.6/-& 100 /98.2/100& 100/98.8/100
 \\
 \text{Leather}& 98.7/97.3/-& 100/99.2/-& 93.9/96.9/-& 100/98.8/-& 99.8/98.8/-& 100/99.6/100& 100/99.7/100
 \\
 \text{Tile}& 99.8/98.0/-& 96.0/90.3/-& 93.7/93.1/-& 99.3/91.8/-& 96.8/92.4/-& 99.5/97.0/98.3 & 99.9/99.0/98.3
 \\
 \text{Wood}& 99.8/96.0/-& 93.8/90.8/-& 95.2/84.8/-& 98.6/93.2/-& 99.7/93.3/-& 98.8/90.9/94.9 & 99.8/97.9/92.4
 \\
 \hline
Texture avg.& 99.1/97.7/-& 95.6/95.4/-&85.7/82.8/-& 99.2/95.8/-& 98.8/95.9/- &\underline{99.6}/\underline{97.0}/\textbf{98.3}& \textbf{99.9}/\textbf{99.0}/\underline{98.1}
\\
\hline
Total avg.& 88.1/87.2/-& 96.4/95.7/-& 85.1/88.9/-& 96.5/\underline{96.8}/-& 97.2/96.8/-& \underline{97.4}/93.1/\underline{93.3}&  \textbf{97.7}/\textbf{97.3}/\textbf{94.8}
\\
\hline

\hline
\end{tabular}}
\label{unsp}
\end{center}
\vspace{-10pt}
\end{table*}
The loss function incorporates a modulating factor ($\gamma$) that tunes down the effect of well-classified instances on the total loss,alongside an optional balancing factor ($\alpha$) to further adjust for the disparity between classes. By doing so, Focal Loss enhances the model's recall on minority classes while maintaining precision.

\begin{center}
\vspace{-10pt}
\begin{equation}
L_{f} = -\frac{1}{n}\sum_{i=1}^{n}\alpha_t (1-p_t)^{\gamma}log(p_t)
\end{equation}
\vspace{-10pt}
\end{center}
where $n = H \times W$ represents the total number of pixels,$p_t$ is the probability of belonging to the true category predicted by the model. In this paper, $p_t$ is the probability of being predicted as an anomaly.

\textbf{Dice Loss} is a performance metric turned loss function widely used in segmentation tasks to evaluate and optimize the overlap between the predicted segmentation mask and the ground truth. It measures the similarity between two samples by calculating the ratio of twice the area of intersection to the sum of the areas of the two samples. The function is particularly effective in scenarios with class imbalance due to its focus on the proportion of correctly predicted pixels relative to the total number of pixels in the target class. By minimizing DICE Loss during training, models are encouraged to produce segmentation outputs that have high spatial overlap with the true object boundaries, making it especially valuable for medical image analysis and other applications requiring precise boundary delineation.
\begin{center}
\vspace{-10pt}
\begin{equation}
L_{d} = -\frac{2\sum_{i=1}^{n} p_t(1-p_t)}{\sum_{i=1}^{n} p_t^2 + \sum_{i=1}^{n} (1-p_t)^2}
\end{equation}
% \vspace{-10pt}
\end{center}
where $p_t$ represents the probability of being predicted as an anomaly.

\begin{center}
\vspace{-10pt}
\begin{equation}
L_{total} = {\lambda}_c \cdot L_c + {\lambda}_f \cdot L_f + {\lambda}_d \cdot L_d,
\end{equation}
% \vspace{-10pt}
\end{center}
% where ${\lambda}_f$ and ${\lambda}_d$ are set to 1 for stage 2 to supervise the training of the decoder. In all other stages, these two coefficients are set to 0. And we utilize the cross-entropy loss throughout various stages of training, ${\lambda}_c$ is set to 1 for all stages. This staged learning strategy allows the model to focus on different components of the loss at different stages, facilitating a more stable and effective training process. This training strategy is further illustrated in Fig.~\ref{fig:training}.
where $\lambda_f$ and $\lambda_d$ are set to 1 in stage 2 to supervise the training of the decoder. In all other stages, these coefficients are set to 0. In contrast, the cross-entropy loss is utilized throughout all training stages, and accordingly, $\lambda_c$ is set to 1 across all stages. This staged learning strategy enables the model to focus on different components of the loss function at each stage, leading to a more stable and effective training process. Further details of this training protocol are illustrated in Fig.~\ref{fig:training}.

\section{Experiment}
\subsection{Datasets}
Our experiments are based on MVTec-AD \cite{mvtecad} and VisA \cite{visa} datasets. Both benchmarks have diverse subsets of different objects, e.g., capsules, leather. In the realm of IAD, MVTec-AD stands out as a widely recognized benchmark. It contains 15 distinct categories, with a total of 3,629 training images and 1,725 testing images. The images within this dataset exhibit resolutions ranging from 700x700 to 1024x1024 pixels, offering a diverse array of visual data for model training.
The recently introduced VisA dataset adds to the resources available for IAD research. Spanning 12 categories, it features 9,621 normal images and 1,200 anomalous images, with an approximate resolution of 1500x1000 pixels.

Following previous IAD methodologies, only the normal data from these two datasets are utilized during the training phase. To address the limitation of insufficient anomalous data and enable effective model training, synthetic anomalous images are generated and incorporated into the training process.

\begin{table*}[htbp]
\caption{
% Few-shot IAD results on MVTec-AD and VisA datasets. Results are listed as the average of 5 runs and the best-performing method is in bold. The results for SPADE, PatchCore and WinCLIP are reported from \cite{jeong2023winclip}.
FEW-SHOT IAD RESULTS ON MVTEC-AD AND VISA DATASETS. RESULTS ARE LISTED AS THE AVERAGE OF 5 RUNS AND THE BEST-PERFORMING METHOD IS IN \textbf{BOLD}. THE RESULTS FOR SPADE, PATCHCORE AND WINCLIP ARE REPORTED FROM \cite{jeong2023winclip}.
}
\vspace{-5pt}

\begin{center}
\setlength{\tabcolsep}{3.8mm}{
\begin{tabular}{c c c c c c c c}
\hline

\hline
\text{Setup}&\text{Method}&\multicolumn{3}{c}{\text{MVTec-AD}}&\multicolumn{3}{c}{\text{VisA}} \\
\cline{3-8} 
 & & \text{I-AUROC} &\text{P-AUROC} & \text{Accuracy} &\text{I-AUROC} & \text{P-AUROC} &\text{Accuracy}\\
\hline

&SPADE \cite{spade} & 81.0±2.0 &91.2±0.4 & -&79.5±4.0 & 95.6±0.4 &-\\
 % & PaDiM & 76.6 ± 3.1 & 89.3 ± 0.9 & - &  62.8 ± 5.4 &  89.9 ± 0.8 & -\\
 &  PatchCore \cite{patchcore} & 83.4±3.0 & 92.0±1.0 & - & 79.9±2.9 & 95.4±0.6 & -\\
1-shot & WinCLIP \cite{jeong2023winclip} & 93.1±2.0 & 95.2±0.5 & - & 83.8±4.0 & \textbf{96.4±0.4} & -\\
 &AnomalyGPT \cite{gu2024anomalygpt} & \textbf{94.1±1.1} & \textbf{95.3±0.1} & 86.1±1.1 & \textbf{87.4±0.8} & 96.2±0.1 & 77.4±1.0 \\
\cline{2-8} 
 &\textbf{IAD-GPT (Ours)} & \textbf{94.1±1.1} & \textbf{95.3±0.1} &\textbf{89.5±1.2} & \textbf{87.4±0.8} & 96.2±0.1 & \textbf{79.1±0.9} \\
\hline

%%%%%%%%%%%%%%%%%%%%%%%%%%
 & SPADE \cite{spade}& 82.9±2.6 & 92.0±0.3 & - & 80.7±5.0 & 96.2±0.4 & - \\
 % & PaDiM & 76.6 ± 3.1 & 89.3 ± 0.9 & - &  62.8 ± 5.4 &  89.9 ± 0.8 & -\\
 & PatchCore \cite{patchcore}&  86.3±3.3 & 93.3±0.6 & - & 81.6±4.0 & 96.1±0.5 & -\\

2-shot & WinCLIP \cite{jeong2023winclip}& 94.4±1.3 & \textbf{96.0±0.3} & - & 84.6±2.4 &  \textbf{96.8±0.3} &  -\\
 &AnomalyGPT \cite{gu2024anomalygpt}& \textbf{95.5±0.8} & 95.6±0.2 & 84.8±0.8 & \textbf{88.6±0.7} &  96.4±0.1 & 77.5±0.3 \\
\cline{2-8} 
 & \textbf{IAD-GPT (Ours)} & \textbf{95.5±0.8} & 95.6±0.2
 &  \textbf{87.7±1.2} &  \textbf{88.6±0.7} &  96.4±0.1 &  \textbf{78.9±0.8}  \\
\hline
%%%%%%%%%%%%%%%%%%%%%%%%%%
 & SPADE \cite{spade}&  84.8±2.5 &  92.7±0.3 &  -  & 81.7±3.4 &  96.6±0.3 &  -\\
 % & PaDiM & 76.6 ± 3.1 & 89.3 ± 0.9 & - &  62.8 ± 5.4 &  89.9 ± 0.8 & -\\
 &  PatchCore \cite{patchcore}&  88.8±2.6 &  94.3±0.5 &  - &  85.3±2.1 &  96.8±0.3 &  -\\
4-shot & WinCLIP \cite{jeong2023winclip}& 95.2±1.3 & 96.2±0.3  & - & 87.3±1.8 & \textbf{97.2±0.2} & -\\
 &AnomalyGPT \cite{gu2024anomalygpt}& \textbf{96.3±0.3} &  \textbf{96.2±0.1} &  \textbf{85.0±0.3} &  \textbf{90.6±0.7} &  96.7±0.1 &  77.7±0.4 \\
\cline{2-8} 

 &\textbf{IAD-GPT (Ours)} & \textbf{96.3±0.3} &  \textbf{96.2±0.1} & 84.0±0.5 & \textbf{90.6±0.7} &  96.7±0.1 & \textbf{78.5±0.5} \\
\hline

\hline

\end{tabular}}
\label{few-shot}
\end{center}
\vspace{-10pt}
\end{table*}

\subsection{Evaluation Metrics}
Following traditional IAD methods, we employ the Area Under the Receiver Operating Characteristic (AUROC) as our evaluation metric for both detection and localization, which is expressed as Image-level AUROC (I-AUROC) and Pixel-level AUROC (P-AUROC).
With the deployment of LLM, existing methods allow determining the presence of anomalies without the need to manually set thresholds. We utilize image-level accuracy to evaluate the performance of our IAD-GPT.

\subsection{Implementation Details}
We use ImageBind-Huge \cite{imagebind} as a frozen image encoder to extract image features and Vicuna-7B \cite{vicuna} as LLM for reasoning, connect them through with TGE. Then We initialize our IAD-GPT using pre-trained parameters from PandaGPT \cite{pandagpt}.
% We use ImageBind-Huge as a frozen image encoder to extract image features and Vicuna-7B as LLM for inference. and connect them using a simple linear layer at first.
We layered the training into three stages, which are stage one to train TGE, stage two to train Visual-guided decoder and MMF, and stage three to train TGE and MMF jointly. At different training stages, we used the same 50 epochs on two V100 GPUs with a learning rate of 0.0005 and a batch size of 16.

Our training strategy is shown in Fig.~\ref{fig:training}. In the first stage, we do not input the mask information generated by the expert model but only train the model to better recognize the anomalous features at the image level.
In the second stage, we freeze TGE on the basis of the first stage, and then train Visual-guided docoder and MMF, which initially aligns the pixel-level anomalous features of the mask to the feature space of the LLM. Finally, in the third stage we freeze the Visual-guided decoder and jointly train TGE and MMF to achieve a better understanding of image-level and pixel-level anomalies in the LLM.

We initialize the image as $224 \times 224$ and similar to AnomalyGPT \cite{gu2024anomalygpt}, without specifying a particular level select the intermediate features of the 8th, 16th, 24th, and 32th layers from the image encoder as input to the decoder. Linear warm-up and a one-cycle cosine learning rate decay strategy are applied.
For image augmentation, the NSA \cite{NSA} method is adopted, with key parameters configured as follows: Poisson image editing is implemented in normal clone mode to achieve smooth edge fusion between synthetic anomalous patches and the original image background; pixel values at the edges of patch masks are set to zero to suppress visible fusion artifacts; and the fusion center is defined as the geometric midpoint of the target pasting region in the destination image, ensuring alignment between the anomalous patch and surrounding image content.
We perform alternating training using both the pre-training data of PandaGPT and our anomaly image-text data. Only TGE, Visual-guided docoder, and MMF perform parameter updates at the corresponding stage, while the rest of the parameters remain frozen.

\begin{figure*}
\centering
%\vspace{-0.8em}
% \includegraphics[width=8.0cm,height=6.2cm]{img/network.pdf}
\includegraphics[scale=0.58]{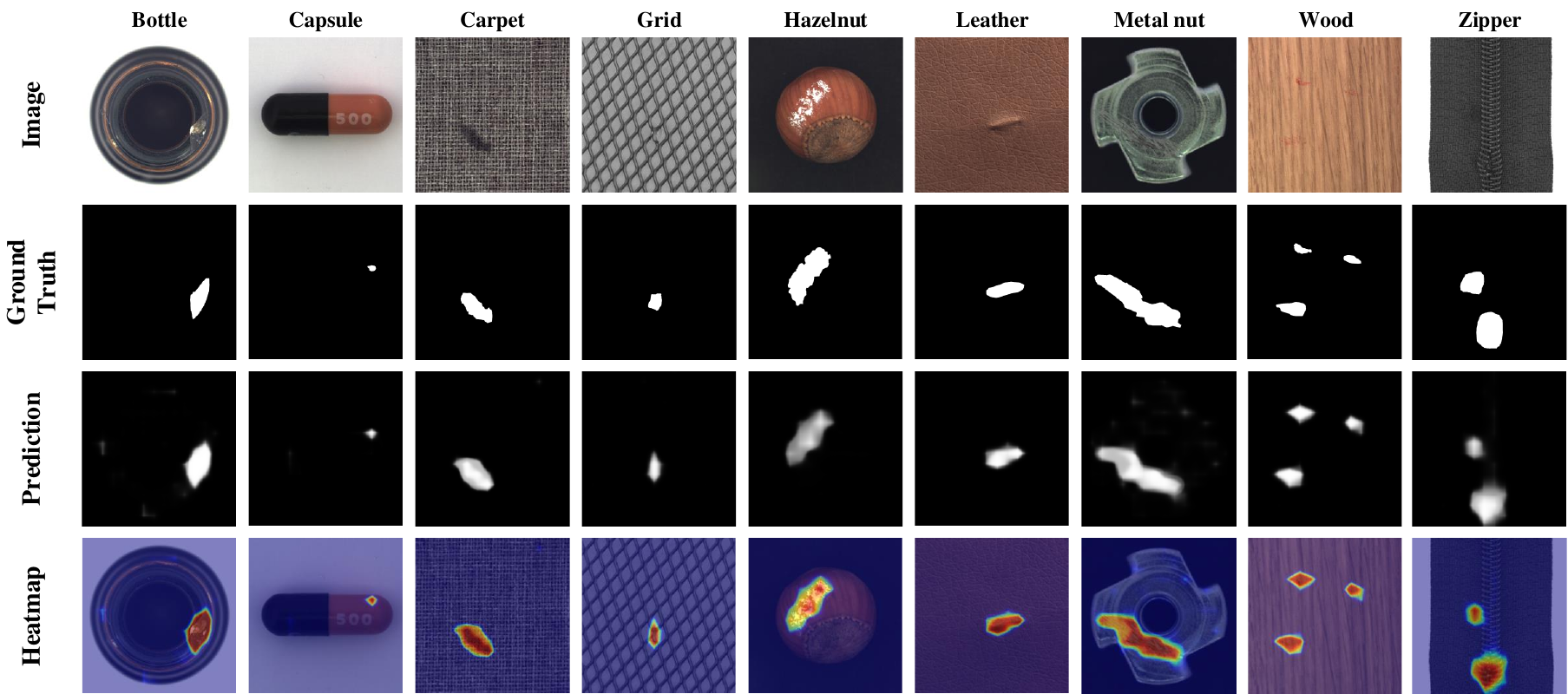}
% \vspace{-1.8em}
\caption{
  % Qualitative evaluation of IAD-GPT on MVTec-AD.
  Qualitative evaluation of IAD-GPT on MVTec-AD. The first row shows the input images from different categories, the second row presents the corresponding ground truth annotations, the third row displays the anomaly detection results predicted by IAD-GPT, and the fourth row visualizes the prediction results using heatmaps.
  }
\vspace{-5pt}
% \caption{\parbox{\textwidth}{\raggedright Overview of IAD-GPT}}
\label{fig:visual}
\end{figure*}
\vspace{-4pt}
\subsection{Self-supervised Industrial Anomaly Detection}
% \textbf{Unsupervised Industial Anomaly Detection.}
In the setting of self-supervised training with a large number of normal samples, given that our method trains a single model on samples from all classes within a dataset, we selected AnomalyGPT \cite{gu2024anomalygpt}, which is trained under the same setup, as a baseline for comparison. Additionally, we compare our model with Draem \cite{zavrtanik2021draem}, PatchCore \cite{patchcore}, SimpleNet \cite{simplenet}, UniAD \cite{uniad} and DiAD \cite{he2024diffusion} using the same unified setting. The results in the MVTec-AD dataset are presented in Table~\ref{unsp}.
Our proposed method, IAD-GPT, demonstrates superior performance compared to existing methods in most categories. 
We have achieved state-of-the-art performance across multiple metrics. For Image-AUROC and Pixel-AUROC, we achieve improvements of $0.3\%$ and $4.2\%$, respectively, compared to AnomalyGPT.
In the task of anomaly segmentation, we demonstrated a significant improvement over AnomalyGPT, demonstrating that APG is effective in promoting large pre-trained models to perceive anomalous features at the patch level. Among multiple multi-category anomaly detection models, our anomaly detection and localization capabilities are the best, reaching $97.7\%$ and $97.3\%$. In the task of APMLLM, our accuracy rate reaches $94.8\%$, representing a relative improvement of $1.5\%$ compared to AnomalyGPT.

\begin{figure}
\centering
%\vspace{-0.8em}
% \includegraphics[width=8.0cm,height=6.2cm]{img/network.pdf}
\includegraphics[scale=0.7]{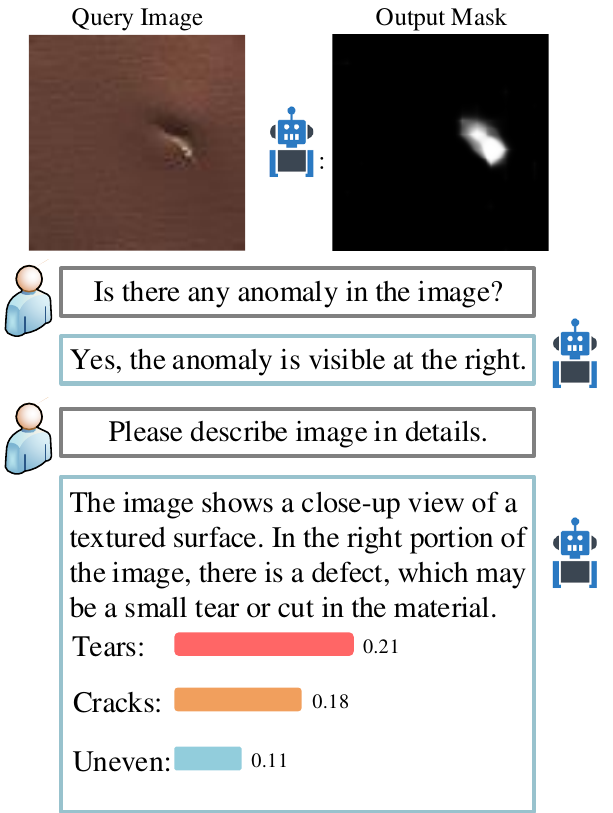}
\caption{
  Qualitative example of IAD-GPT on MVTec-AD. Anomaly categories are computing from the similarity between $F_{img}$ and $F_{apg}$.}
\vspace{-10pt}
\label{fig:demo}
\end{figure}

\subsection{Few-shot Industrial Anomaly Detection}
We compare our work with prior few-shot IAD methods, selecting SPADE \cite{spade}, PatchCore \cite{patchcore}, WinCLIP \cite{jeong2023winclip}, and  AnomalyGPT as the baselines. The results are presented in Table \ref{few-shot}. Across both datasets, Our method performs competitively in the IAD and APMLLM tasks, and notably outperforms AnomalyGPT in the setting of 1-shot and 2-shot and achieves state-of-the-art performance. 
\begin{table}[htbp]
\vspace{15pt}
\setlength{\tabcolsep}{2pt}
\caption{
\MakeUppercase{Ablation of TGE in different frameworks. Acc. denotes Accuracy (\%).}
}
\begin{center}
\begin{tabular}{c c c c c c}
\hline
TGE & IAD-GPT  & AnomalyGPT& I-AUROC & P-AUROC & Acc.\\
\hline
& & & - & - & 72.2 \\
\checkmark& &  & - & - & 82.3 \\
% & \checkmark & & 97.5 & 95.6 & 94.0 \\
& &\checkmark & 97.3 & 93.1 & 93.3 \\
\checkmark& &\checkmark & 97.3 & 93.1 & 93.6 \\
\checkmark & \checkmark & & \textbf{97.7} & \textbf{97.3} & \textbf{94.0} \\
\hline
\end{tabular}
\label{AblationTGE}
\end{center}
\vspace{-15pt}
\end{table}
Compared to AnomalyGPT, our method achieves better performance on Mvtec-AD and VisA for most metrics. In the 1-shot and 2-shot setting of the Mvtec-AD, the accuracy of IAD-GPT is $89.5\pm1.2 \%$ and $79.1\pm0.9 \%$, which is improves by $3.4\%$ and $1.7\%$ over AnomalyGPT. 
In other settings, IAD-GPT also achieves competitive results. This indicates that our multi-scale feature enhancement approach effectively improves the LLM's ability to perceive anomalies.

In the few-shot in-context learning setting, the localization performance of the model is slightly lower than that of the self-supervised setting due to limited normal references. Our proposed use of TGE and MMF to provide multi-scale anomaly perception for LLMs, which promotes the performance of LLMs in the APMLLM task. Notably, AnomalyGPT exhibits weaker anomaly localization capabilities in a self-supervised setting compared to the abilities of the model in a few-shot learning setting without training.
This indicates that AnomalyGPT does not fully leverage the capabilities of large pre-trained models. However, our proposed APG effectively compensates for this shortcoming. IAD-GPT achieves an anomaly localization performance of $97.3\%$ P-AUROC in the self-supervised setting, surpassing the best result of $96.2\%$ in the few-shot setting.
\begin{table}[htbp]
\setlength{\tabcolsep}{3pt}
\caption{
\MakeUppercase{
Ablation study on the integration of expert knowledge into LLM.
}}
\begin{center}
\begin{tabular}{c c c c c c}
\hline
APG & MMF  & Prompt Learner & I-AUROC & P-AUROC & Acc.\\
\hline
& &\checkmark & 97.3 & 93.1 & 93.3 \\
\checkmark& &\checkmark & 97.5 & 95.6 & 93.6 \\
\checkmark & \checkmark & & \textbf{97.7} & \textbf{97.3} & \textbf{94.0} \\
\hline
\end{tabular}
\label{AblationAPG}
\end{center}
\vspace{-10pt}
\end{table}
\begin{table}[htbp]
\setlength{\tabcolsep}{3pt}
\caption{\MakeUppercase{
Comparison of Prompt Learner from AnomalyGPT and MMF from IAD-GPT on Throughput (imgs/s), Parameters (M), FLOPs (G), and Accuracy (\%).
}}
% \vspace{-10pt}
\begin{center}
\begin{tabular}{c c c c c}
\hline
Module & Throughput$\uparrow$ & Parameters$\downarrow$ & FLOPs$\downarrow$ & Acc.$\uparrow$\\
\hline
Prompt Learner & 97.8 & 107.4M & \textbf{15.6G} & 93.3 \\
MMF & \textbf{114.2} & \textbf{10.2M} & 49.3G & \textbf{94.8} \\
\hline
\end{tabular}
\label{AblationMMF}
\end{center}
\vspace{-10pt}
\end{table}
\subsection{Qualitative Examples}
The visualization results of IAD-GPT on the MVTec-AD dataset can be seen in Fig.~\ref{fig:visual}.
% We provide query images and ground textures in the first and second rows of the figure. The third row shows the prediction of anomalies by IAD-GPT, and the fourth row presents the perception of anomalous regions by IAD-GPT in the form of a heatmap. From the Fig.~\ref{fig:visual}, it
It can be seen that IAD-GPT effectively identifies anomalies of different categories and has good perceptual ability in pixel-level anomaly localization.
Regardless of the scale of the anomaly, whether it be large scratches or small pokes, IAD-GPT demonstrates high accuracy in both detection and localization.
Fig.~\ref{fig:demo} illustrates the performance of our IAD-GPT in self-supervised anomaly detection. Our model can not only indicate the existence of anomalies, accurately locate their locations, and provide pixel-level localization results, but also answer specific categories of anomalies that may exist, which is a capability that AnomalyGPT does not possess. Users can engage in multi-turn conversations related to the image content, including but not limited to asking IAD-GPT whether the image contains anomalies or requesting specific descriptions about the image.

\subsection{Ablation Study}
To evaluate the effectiveness of each proposed module, extensive ablation experiments were conducted on the MVTec-AD dataset. Our study primarily focuses on three key aspects: the Text-Guided Enhancer, the integration of expert knowledge, and the multistage training strategy. 
The main results are summarized in Table~\ref{AblationTGE},~\ref{AblationAPG},~\ref{AblationMMF} ,~\ref{AblationTRAIN} and~\ref{AblationDataAug}.
All analyses are based on self-supervised training and testing protocols applied to the MVTec-AD dataset.
\subsubsection{Impact of TGE}
To demonstrate the effectiveness of the TGE in enhancing visual information, we train the model for anomaly perception using only the TGE. As shown in Table~\ref{AblationTGE}, compared to PandaGPT, our approach achieves a performance improvement of $10.1\%$.
To further validate the applicability of the TGE across different frameworks, we also conducted ablation studies on AnomalyGPT. 
The experimental results confirm that the TGE consistently improves the model’s ability to perceive anomalies at the image level, thereby enabling better performance on APMLLM task in both IAD-GPT and AnomalyGPT.

\subsubsection{Impact of Expert Knowledge}
\begin{table}[htbp]
\setlength{\tabcolsep}{3pt}
\caption{
\MakeUppercase{
Ablation of train strategy.
}}
\begin{center}
\begin{tabular}{c c c c c c}
\hline
Multi- & IAD-GPT  & Anomaly- & I-AUROC & P-AUROC & Acc.\\
stage &  & GPT &  &  & \\
\hline
& &\checkmark & 97.3 & 93.1 & 93.3 \\
% \checkmark &  & \checkmark & 97.3 & 93.3 & 93.6 \\
& \checkmark & & \textbf{97.7} & \textbf{97.3} & 94.0 \\
\checkmark & \checkmark & & \textbf{97.7} & \textbf{97.3} & \textbf{94.8} \\
\hline
\end{tabular}
\label{AblationTRAIN}
\end{center}
\vspace{-10pt}
\end{table}
\begin{table}[htbp]
\setlength{\tabcolsep}{10pt}
\caption{
\MakeUppercase{
Comparison of Image Augmentation Methods.
}}
\begin{center}
\begin{tabular}{c c c}
\hline
Method & I-AUROC  & P-AUROC\\
\hline
NSA~\cite{NSA} & \textbf{97.7} & \textbf{97.3} \\
CutPaste~\cite{li2021cutpaste} & 92.1 & 89.2 \\
NSA~\cite{NSA} + CutPaste~\cite{li2021cutpaste} & 94.1 & 91.1 \\
\hline
\end{tabular}
\label{AblationDataAug}
\end{center}
\vspace{-10pt}
\end{table}
To demonstrate the impact of the expert knowledge incorporated via APG and MMF, we compare the performance of AnomalyGPT with our method after integrating expert knowledge. As shown in Table~\ref{AblationAPG}, APG consistently improves both anomaly detection and localization across different frameworks, indicating that APG is effective in promoting large pre-trained models to perceive anomalous features at the patch level.

To enable the LLM to better comprehend and utilize the expert knowledge, we propose the MMF module. Unlike the Prompt Learner used in AnomalyGPT, MMF fully exploits multi-level expert knowledge during the prompting process. In Table~\ref{AblationMMF}, we compare the efficiency of MMF and the Prompt Learner. MMF achieves superior performance in terms of throughput and parameter count, reaching 114.2 imgs/s and 10.2M parameters, compared to 97.8 imgs/s and 107.4M parameters for the Prompt Learner. However, due to the additional overhead of processing multi-layer expert knowledge, MMF incurs a higher computational cost, as reflected by its significantly larger FLOPs. In terms of accuracy, IAD-GPT achieves $94.8\%$, outperforming AnomalyGPT’s $93.3\%$, demonstrating the effectiveness of our design.

\subsubsection{Impact of Training Strategy}
To evaluate the effectiveness of the multi-stage training strategy, we present its impact on IAD-GPT in Table~\ref{AblationTRAIN}. Without multi-stage training, our method still outperforms AnomalyGPT across all evaluation metrics. 
The incorporation of multi-stage training further enhances IAD-GPT’s ability to perceive anomalies in APMLLM task, leading to improved performance in both detection and localization.

{
\subsubsection{Impact of Data Augmentation}
We have supplemented a more detailed comparative ablation experiment focusing on data augmentation methods, specifically evaluating three scenarios: training with only the NSA-based augmentation, training with only the CutPaste augmentation, and training using a combination of NSA and CutPaste. All experiments strictly followed the experimental setup in IV-C. For the combination scheme, we randomly selected either NSA or CutPaste to synthesize anomalous images in each training iteration before feeding them into the model.

The experimental results in Table~\ref{AblationDataAug} show that the NSA-based augmentation achieves the best performance in both anomaly detection and localization tasks. We believe this is attributed to its ability to generate more realistic anomalous regions with smoother edge transitions, which helps the model learn more discriminative normal-abnormal feature differences.
}
\section{Conclusion}
In this study, we introduce IAD-GPT, an innovative framework for IAD.
IAD-GPT leverages the advanced capabilities of MLLMs and integrates multi-scale visual information through TGE and MMF. 
TGE effectively enhances the alignment between image-level visual information and LLMs, and improves LLM's perception of anomalies by dynamically selecting enhancement paths for image features.
Meanwhile, the MMF module integrates multi-level localization results as visual expert knowledge for LLM to enhance its pixel-level anomaly perception.
Our experiments on benchmark datasets such as MVTec-AD and VisA highlight the superior performance of IAD-GPT.
IAD-GPT achieves better performance in APMLLM task by leveraging multi-scale visual information. Furthermore, it fully enhances the capabilities of large pre-trained models based on APG to detect and localize image anomalies.
We have improved our performance in anomaly detection and localization compared to the baseline, and due to the excellent performance of APG, we have achieved better anomaly localization performance in the self-supervised setting than in the few-shot in-context learning setting.

IAD-GPT provides a more comprehensive and robust LLM-based solution for industrial applications. Beyond its technical contributions, this work also underscores the broader potential of leveraging MLLMs in industrial domains, opening new avenues for interactive and explainable artificial intelligence solutions.
Future work will explore the extension of IAD-GPT to other fields, such as medical anomaly detection and camouflage object detection. In addition, efforts will be made to improve its adaptability to more complex industrial scenarios.

\ifCLASSOPTIONcaptionsoff
  \newpage
\fi

\bibliographystyle{IEEEtran}
\bibliography{IEEEabrv,reference}

%\begin{IEEEbiography}[{\includegraphics[width=1in,height=1.25in,clip,keepaspectratio]{Figures/Zitong.png}}]{Zitong Yu}
% received the M.S. degree from University of Nantes, France, in 2016, and he is currently a Ph.D. candidate in the Center for Machine Vision and Signal Analysis, University of Oulu, Finland. His research interests focus on remote physiological measurement, face anti-spoofing and video understanding. He led the team and won the 1st Place in the ChaLearn multi-modal face anti-spoofing attack detection challenge with CVPR 2020.
%\end{IEEEbiography}

\end{document}